\documentclass[letterpaper, 10 pt, journal, twoside]{IEEEtran}
\usepackage{needspace} % you already have it
\newcommand{\TopTextCushion}{\Needspace{1\baselineskip}\vspace*{0.1ex}}


\usepackage{caption}
\usepackage{needspace}

\usepackage{graphics}
\usepackage{amsmath}
\usepackage{amsfonts}
\usepackage{colortbl}
\usepackage{amssymb} % to use triangleq
\usepackage{graphicx}
\usepackage{enumerate}   
\makeatletter
\let\NAT@parse\undefined
\makeatother
\usepackage{hyperref}
\usepackage[sort,compress]{cite}

\usepackage{lipsum}

\usepackage{balance}
\usepackage{arydshln}

\usepackage{booktabs}
\usepackage{tikz}

\usepackage{caption}
\usepackage{mathrsfs}

\usepackage[captionskip=2pt]{floatrow}
\newlength\Myfigwd
\usepackage{tabularx}
\newcommand{\realfield}{\hbox{I \kern -.4em R}}
\newcommand {\mb}[1]{\mathbf{#1}} % all replaced
\newcommand {\bs}[1]{\boldsymbol{#1}}

\newcommand{\T}{^{\top}}  %shortcut for transpose
\newcommand*{\diameter}{\bigcirc\kern-0.95em\diagup}
\newcommand{\rmd}{\textrm{d}}  %shortcut for derivative
\makeatletter
\newcommand{\rom}[1]{\uppercase\expandafter{\romannumeral #1\relax}}
\makeatother

\usepackage{color}  %needed for color text

\usepackage{caption} % Add this package
\usepackage{tikz}
\usepackage{etoolbox}

\usepackage{multirow}
\usepackage{soul}
\usepackage{xcolor}
\usepackage{cancel} 
\usepackage[normalem]{ulem} 

\newcommand{\removecomment}[1]{}
\newcommand{\editcomment}[2]{{#1}}

\newcommand{\citationremove}[1]{}

\newtoggle{draft}
\usepackage{tikz}
\usetikzlibrary{arrows.meta,positioning,fit,calc,decorations.pathmorphing}

\graphicspath{{figures/pdf/}}

\begin{document}
% \title{Integrated Shape and Force Sensing for Continuum Instruments via Polynomial Curvature Modeling and Torque-Cell-Based Actuation}
% \title{Torque-Cell-Enabled Actuation Unit and Integrated Shape–Force Estimation for Cable-Driven Continuum Robots}
\title{\fontsize{20pt}{20pt}\selectfont Integrated Shape–Force Estimation for Continuum Robots: A Virtual‑Work and Polynomial‑Curvature Framework}
\author{Guoqing~Zhang\textsuperscript{ 1,\dag}, Zihan~Chen\textsuperscript{ 2,\dag}, and Long~Wang\textsuperscript{ 1,*}% <-this % stops a space
\thanks{${}^{\dagger}$ G. Zhang and Z. Chen contributed equally to this work.}
\thanks{${}^1$ G. Zhang, and L. Wang are with the Department
of Mechanical Engineering, Stevens Institute of Technology, NJ, USA, e-mail: {\tt \{gzhang21, lwang4\}@stevens.edu}, \textsuperscript{*}Corresponding author.}% <-this % stops a space
\thanks{${}^2$ Z. Chen is currently with Cornerstone Robotics Limited, Hong Kong. His contributions to this paper were made before his affiliation with the company. e-mail: {\tt tom.chen@csrbtx.com}}
}

% The paper headers
\markboth{\tiny Manuscript Submitted to 2026 TMECH/AIM Focused Section}%
{\footnotesize Zhang \MakeLowercase{\textit{et al.}}: Polynomial Curvature–Based Shape–Force Estimation}
% {Shell \MakeLowercase{\textit{et al.}}: Bare Demo of IEEEtran.cls for Journals}

\maketitle
\begin{abstract}
Cable-driven continuum robots (CDCRs) are widely used in surgical and inspection tasks that require dexterous manipulation in confined spaces. 
Existing model-based estimation methods either assume constant curvature or rely on geometry-space interpolants, both of which struggle with accuracy under large deformations and sparse sensing.
This letter introduces an integrated shape–force estimation framework that \removecomment{combines cable-tension measurements with tip-pose data to reconstruct backbone shape and estimate external tip force simultaneously.}{}\editcomment{combines tip-pose with cable-tension within a unified two-stage pipeline: the first stage reconstructs the backbone shape from tip-pose observations, and the second stage estimates the external tip wrench from actuator tensions using the reconstructed shape.}{[R1-m1]} The framework employs polynomial curvature kinematics (PCK) and a virtual-work-based static formulation expressed directly in curvature space, where polynomial modal coefficients serve as generalized coordinates.
The proposed method is validated through Cosserat-rod-based simulations and hardware experiments on a torque-cell-enabled CDCR prototype. 
Results show that the second-order PCK model achieves superior shape and force accuracy, combining a lightweight shape optimization with a closed-form, iteration-free force estimation, offering a compact and robust alternative to prior constant-curvature and geometry-space approaches.

\end{abstract}
% % Note that keywords are not normally used for peerreview papers.
% \begin{IEEEkeywords}
% Continuum robots, shape estimation, force estimation, polynomial curvature kinematics, virtual work.
% \end{IEEEkeywords}
\section{Introduction}
\label{sec:introduction}
Cable-driven continuum robots (CDCRs) have emerged as a promising class of manipulators capable of performing dexterous operations in constrained and cluttered environments, such as minimally invasive surgical sites\cite{simaan2004high,wu2022robotic,goldman2012design,huang2024optimal}.
Their long, slender, and continuously deformable backbones allow smooth navigation around obstacles and precise interaction with delicate tissue.
To achieve reliable manipulation in such settings, accurate knowledge of both the robot’s backbone shape and the external forces acting at its distal end is essential for feedback control, motion planning, and safety assurance.
However, directly sensing these quantities remains difficult due to miniaturization limits, sterilization requirements, and incompatibility with medical imaging modalities.
As a result, there has been a growing trend toward model-based estimation techniques that infer shape and force information from sparse, indirect measurements such as tip-pose and actuation tensions.\par
Existing model-based estimation approaches can be grouped into three main families.
\textbf{(i) Shape estimation:} Curve-based models generalize the constant-curvature (CC) assumption to variable-curvature forms that describe more complex geometries such as S-bends or gradual curvature transitions.
Representative approaches include polynomial curvature kinematics (PCK)\cite{della2019control,cheng2022orientation,11268969}, Euler curves and Euler arc splines\cite{rao2021using,9804779}, and other modal bases such as Chebyshev polynomials\cite{orekhov2023lie}. A recent overview provides a broader summary of reduced-order modeling for continuum manipulators\cite{sadati2023reduced}.
Reduced-order segmentation has also been employed in FBG-based strain sensing to refine curvature measurements before global reconstruction\cite{al2021fbg}.
Alternatively, distributed-state estimators that fuse Cosserat-rod priors with Gaussian-process regression achieve continuous shape reconstruction from discrete pose or strain observations\cite{lilge2022continuum}.
Geometry-space Bézier or B-spline curve fits are also common due to their smooth interpolation and simple parameterization\cite{song2015real,yuan2017shape}.
\textbf{(ii) Force estimation:} Intrinsic sensing (actuation-based) methods use static equilibrium and virtual work to relate actuator efforts to the distal wrench, providing physical insight into the robot’s kinetostatics and observability\cite{xu2008investigation,xu2010intrinsic,khoshnam2015modeling,back2015catheter}.
\textbf{(iii) Integrated shape–force estimation:} Several recent works combine the two processes.
Two-stage pipelines first reconstruct shape and then invert statics for force estimation\cite{yuan2017shape}.
Alternatively, probabilistic joint inference approaches leverage Cosserat-rod modeling priors and Gaussian processes to estimate shape and force simultaneously\cite{ferguson2024unified}.\par
\begin{figure}[!t]
    \centering
    \includegraphics[width=0.98\linewidth]{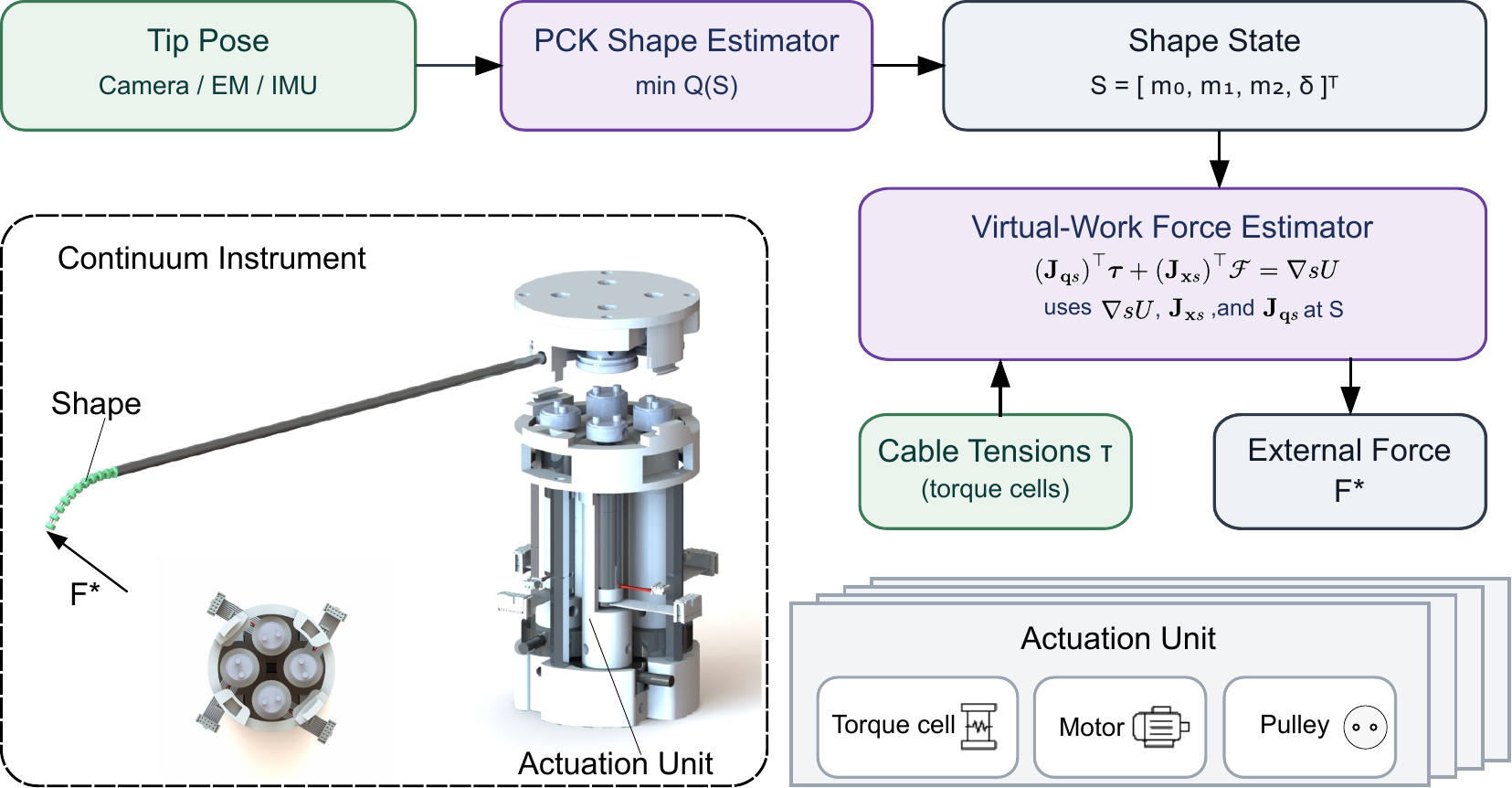}
    \vspace{2mm}
    \caption{Integrated shape–force sensing framework.
Tip pose $\!\rightarrow\!$ PCK shape $\mb{\mathcal{S}}=[m_0,m_1,m_2,\delta]\T$; with tensions $\tau$ and model Jacobians, the virtual-work balance yields $F^{*}$.}
\label{fig:framework-compact}
\end{figure}
While effective within their respective assumptions, each of these families faces notable limitations.
Constant-curvature and geometry-space formulations are computationally light but suffer from large force drift and shape underfitting at high curvatures.
Conversely, distributed-state inference approaches offer high spatial fidelity but rely on high-dimensional, discretization-dependent states. They also require careful calibration and tuning, which can pose challenges for embedded or real-time implementation.
Despite these advances, there remains limited work on reduced-order frameworks that explicitly couple polynomial-curvature shape modeling with virtual-work-based equilibrium for integrated shape--force
estimation from sparse sensing. Such a framework should enable consistent estimation of both shape and external wrenches while maintaining computational tractability.\par
This letter presents an estimator-centric and sensor-agnostic framework for integrated shape–force estimation of cable-driven continuum robots.
The method reconstructs the continuous backbone shape and the external tip wrench from a limited tip-pose sensor configuration and cable-tension measurements. The proposed approach formulates static equilibrium directly in curvature space, where curvature serves as the physically meaningful variable that directly governs bending strain energy and virtual work.
\removecomment{This formulation yields analytic mappings among shape, joint, and task spaces, enabling real-time estimation.}{}\editcomment{This formulation yields explicit mappings among shape, joint, and task spaces, with semi-analytic Jacobians where the remaining 1D integral terms are evaluated numerically, enabling real-time estimation.}{R1-m12}\par
As illustrated in Fig.~\ref{fig:framework-compact}, the framework adopts a two-stage curvature-space pipeline.
\removecomment{The first stage reconstructs curvature using polynomial curvature kinematics (PCK) from sparse tip-pose data.}{}\editcomment{The first stage reconstructs curvature using polynomial curvature kinematics (PCK) from sparse tip-pose measurements, which may be provided by different sensing modalities (e.g., vision-, electromagnetic (EM)-, or inertial measurement unit (IMU)-based sensing).}{R2-C1}
The second stage estimates the external wrench through a virtual-work-based static balance using actuator tensions, resolved via a compact weighted least-squares (WLS) redundancy formulation.
This two-stage factorization ensures modularity, fast computation, and clear separation between sensing and model domains.\par
The proposed formulation addresses the limitations of prior approaches in two aspects.
First, representing the system state using curvature eliminates modeling drift from geometry-space interpolation and improves force stability under large deformation.
Second, the compact polynomial representation captures the continuum’s intrinsic low-dimensional articulation while maintaining high Cartesian fidelity; \removecomment{its closed-form Jacobians and strain-energy gradients enable deterministic, fast updates.}{}\editcomment{its semi-analytic Jacobians and closed-form strain-energy gradients enable deterministic, fast updates.}{R1-m12}
The main contributions of this letter are summarized as follows:
\begin{enumerate}
\item \emph{Curvature-space virtual-work formulation.} 
\removecomment{The backbone mechanics are formulated directly in curvature space, yielding closed-form kinematics and strain-energy gradients that generalize the constant-curvature assumption while remaining analytically tractable.}{}\editcomment{The backbone mechanics are formulated directly in curvature space, yielding semi-analytic Jacobians and closed-form strain-energy gradients that generalize the constant-curvature assumption while remaining computationally efficient.}{R1-m12}
\item \emph{Integrated shape–force pipeline.} 
Polynomial curvature is reconstructed from sparse tip-pose measurements, and the distal wrench is recovered via a virtual-work formulation in curvature space using actuator tensions.
\end{enumerate}\par

Additionally, we introduce a torque-cell-enabled actuation unit
that provides direct, high-bandwidth cable-tension feedback.
Unlike prior designs that inferred tension from motor currents or
in-line force sensors~\cite{jakes2019model,nguyen2024design,chien2023design,grassmann2024open},
each motor integrates a miniature reaction-torque sensor mapped analytically to cable tension through the known wrap geometry,
preserving the native cable routing and compact form factor.\par
The proposed framework is validated through both simulations and hardware experiments.
Planar Cosserat-rod simulations provide ground-truth backbone shapes and tip forces for quantitative evaluation under multiple loading conditions.
Experimental results on a torque-cell-enabled continuum instrument further demonstrate accurate, robust, and consistent estimation performance under realistic constraints.

\section{Estimation framework formulation}
\label{ch: model formulation}
In this section we present a novel virtual-work based modeling framework to support the \removecomment{force-shape}\editcomment{shape-force}{[R1-m3]} estimation function. We define the polynomial curvature-based shape state space and establish the mappings between joint space, shape state space, and task space. Building on these mappings, at the end, we lead to a corresponding force estimation method based on a static model using the virtual-work principle. This method is represented by polynomial curvature kinematics, offering a structured approach to model the continuum instrument's deformation.
\begin{figure}[!t]
    \centering
    \includegraphics[width=.83\columnwidth]{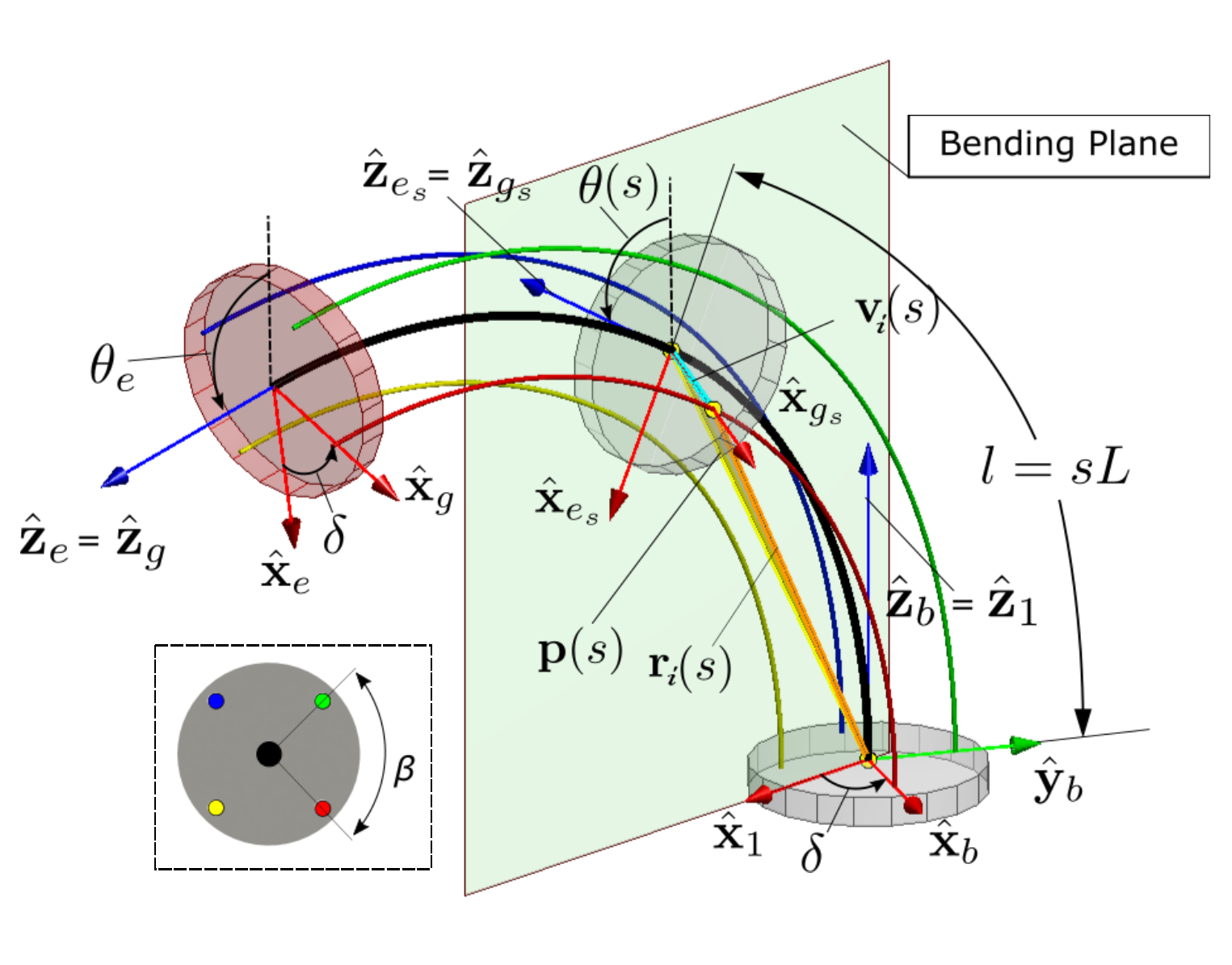}
    \caption{\editcomment{Kinematic frame assignment of the continuum robot. Frame $\{b\}$ is the base frame. Frame $\{1\}$ is obtained from $\{b\}$ by a rotation $-\delta$ about $z_b$, defining the bending plane. Frame $\{e_s\}$ is the local backbone frame at normalized arclength $s$, obtained from frame $\{1\}$ by a rotation $\theta(s)$ about $y_1$; $x_{e_s}$ lies in the bending plane and $z_{e_s}$ aligns with the backbone tangent. Frame $\{g_s\}$ is obtained from $\{e_s\}$ by a rotation $\delta(s)$ about $z_{e_s}$. The tip frames correspond to $s=1$.}{[R1-m5, R1-m8]}}
    \label{fig:sch_kin}
\end{figure}
\subsection{Polynomial Curvature-based Shape Space} \label{ch:pc_kiematics}
Let \(L\) be the physical length of the continuum segment, and let \(l \in [0, L]\) denote the local arclength measured from the instrument’s base. We define the normalized arclength as \(s = \frac{l}{L}\), ensuring \(s \in [0,1]\). We assume that the backbone deforms within a single bending plane, without torsion or out-of-plane deflection. The orientation of this plane may rotate about the base frame depending on the actuation pattern. \editcomment{Here, "planar bending" does not mean that the backbone is restricted to a fixed plane in the workspace. The bending-plane angle $\delta$ is part of the shape state and rotates with the actuation pattern, allowing the robot to bend toward different directions in 3D space. The current model captures this primary bending behavior, while torsion and general out-of-plane deformation are intentionally excluded.}{R1-M2,R2-C2} Unlike the constant curvature assumption, a more precise kinematics-based representation of the shape is polynomial curvature kinematics \cite{wang2019geometric,cheng2022orientation,11268969}. Polynomial curvature kinematics represent the curvature distribution along the arclength of the continuum instrument as a polynomial function. A $j^{th}$ order polynomial model is given by:
\begin{equation}
\label{eqn: bending angle exp}
\kappa(s) \simeq \sum_{i=0}^j m_i s^i, \quad \theta(s)=\int_0^s \kappa(\tau) \mathrm{d}\tau=\sum_{i=0}^{j} m_i \frac{s^{i+1}}{i+1} 
\end{equation}
where $\kappa(s)$ is the curvature at arclength $s$ and $m_i$ is the modal coefficient of $i^{th}$ basis in the polynomial function. The bending angle $\theta(s)$ at the normalized arclength $s$ can be obtained by integrating the above curvature equation.\par
\editcomment{Figure~\ref{fig:sch_kin} illustrates the kinematic frame assignment and transformation structure used to derive the instrument articulation kinematics.
Coordinate frames are identified by their axis subscripts (e.g., $x_b,y_b,z_b$ denote the base frame), and the homogeneous transformation $^{a}\mb{T}_{b}\in SE(3)$ represents the pose of frame $\{b\}$ expressed in frame $\{a\}$.}{[R1-m5]}%
Let $\mb{p}(s)$ be the position at normalized arclength $s$, defined as the task space. The position vector $\mb{p}(s)$ is expressed by the integral of the tangent derivative $\tfrac{\rmd \mb{p}}{\rmd s}$ along the normalized arclength $s$:
\editcomment{\begin{equation}
\label{eqn: position_in_frame1}
\begin{aligned}
   ^{1}\mb{p}(s) & \triangleq \big[p_x(s), p_y(s), p_z(s)\big]\T 
   = \int_0^s  \tfrac{\rmd \mb{p}}{\rmd s} (\xi) \,\rmd \xi\\[0pt]
    & =\int_0^s L \left[\sin\left(\theta(\xi)\right), 0, \cos\left(\theta(\xi)\right)\right]\T \rmd \xi
\end{aligned} 
\end{equation}}{[R1-m4, R1-m11]} \par
Combining with the bending angle $\theta(s)$, the bending direction angle $\delta$ and the position vector $\mb{p}(s)$, the homogeneous transformation at location $s$ can be obtained:
\editcomment{\begin{equation}
{}^{b}\mb{T}_{g_s}(s) = {}^{b}\mb{T}_1 \, {}^{1}\mb{T}_{e_s}(s) \, {}^{e_s}\mb{T}_{g_s}
\end{equation}}{}%
\editcomment{where frame $\{1\}$ is obtained from frame $\{b\}$ by a rotation $-\delta$ about $z_b$, i.e., ${}^{b}\mb{T}_1=\mathrm{diag}\!\left(e^{-\delta[\mb{e}_3^\wedge]},1\right)$.}{[R1-m5]} %
Also, ${}^{e_s}\mb{T}_{g_s}\!=\!\mathrm{diag}\!\left(e^{\delta[\mb{e}_3^\wedge]},1\right)$, and
$${}^{1}\mb{T}_{e_s}(s)=\begin{bmatrix}
e^{\theta(s)[\mb{e}_2^\wedge]} & {}^{1}\mb{p}(s)\\[3pt]
\mb{0} & 1
\end{bmatrix}\!.$$ \editcomment{
The operator $(\cdot)^\wedge$ denotes the skew-symmetric matrix associated with a vector in $\mathbb{R}^3$, satisfying $\mathbf{a}^\wedge \mathbf{b} = \mathbf{a} \times \mathbf{b}$.
}{[R1-m9]}

In this work, we consider the highest order of the curvature polynomial as $j=2$. Thus, by integrating with the bending direction angle $\delta$, we can reconstruct the shape space of the continuum instrument as:
\begin{equation}
    \mb{\mathcal{S}} = [m_0, m_1, m_2, \delta]\T
    \label{eqn: config_def}
\end{equation}
\editcomment{Accordingly, $S$ captures curvature variation within the rotating bending plane only.}{R1-M2,R2-C2}
\subsection{Mapping Between Joint and Shape Spaces}
Figure~\ref{fig:sch_kin} illustrates the inverse kinematics of \textit{joint-to-configuration} space for a continuum segment actuated by four pulling wires. We use $\mb{L} = \left[L_1, L_2, L_3, L_4\right]\T$ to represent the pulling wire lengths. The length of $i^{th}$ pulling wire $L_i$ can be derived using the position vector along its length, $\mb{r}_i(s)$.
\editcomment{\begin{equation}
\label{eqn: L_i def}
\begin{aligned}
    L_i = \int_0^L\left\|\tfrac{d \mb{r}_i(s)}{d l}\right\| d l = \int_0^L\left\|\tfrac{d \mb{r}_i(s)}{d (sL)}\right\| d (sL) = \int_0^1\left\|\tfrac{d \mb{r}_i(s)}{d s}\right\| d s
\end{aligned}
\end{equation}}{[R1-m6]}
The position vector, $\mb{r}_i(s)$, can be expressed as the sum of two vectors: the central backbone position vector $\mb{p}(s)$ and a radial offset vector $\mb{v}_i(s)$, both of which are illustrated in Fig.~\ref{fig:sch_kin}. Therefore we have,
\begin{equation}
  \frac{d \mb{r}_i(s)}{d s}=\frac{d \mb{p}(s)}{d s}+\frac{d \mb{v}_i(s)}{d s}  
\end{equation}
The central backbone tangent derivative $\frac{d \mb{p}(s)}{d s}$ was already obtained from \eqref{eqn: position_in_frame1}, and the radial offset derivative \editcomment{$\frac{d \mb{v}_i}{d s}$}{} can then be derived as:
\editcomment{\begin{align}
\frac{d \mb{v}_i}{d l} & = \frac{d \mb{v}_i}{d (sL)} = { }^1 \mb{R}_{e_s}\left({ }^{e_s} \boldsymbol{\omega}_{g_s} \times{ }^{e_s} \mb{v}_i\right) \label{eqn: dv/ds}\\
{ }^{e_s} \boldsymbol{\omega}_{g_s} & =\mb{e}_2 \tfrac{d \theta(s)}{d s}=\left[0,\kappa(s), 0\right]\T \\[2pt]
{ }^1 \mb{R}_{e_s} & =e^{\theta(s)\left[\mb{e}_2{ }^{\wedge}\right]},\quad { }^{e_s} \mb{v}_i=r\left[\cos(\sigma_i), \sin(\sigma_i), 0\right]\T
\end{align}}{[R1-m4, R1-m6]}
where $\sigma_i(s)$ is the angular coordinate of $i^{th}$ pulling wire with respect to \editcomment{frame $\{e_s\}$}{[R1-m8]}. We define
\begin{equation}
\sigma_i=\delta+(i-1) \beta
\end{equation}
with \(\beta\) being the angular spacing between cables and \(\delta\) the global bending direction angle aligns the instrument’s bending plane, ensuring each cable’s radial offset is rotated accordingly.
Also, \(r\) designates the constant distance between the central backbone and the pulling wires. Then the pulling wire length  $L_i$ is derived as:
\editcomment{\begin{equation}
\label{eqn: L_i expre}
\begin{aligned}
    L_i &= \int_0^1\left\|L-r\kappa(s)\cos(\sigma_i)\right\| d s
        &= L - \theta_er\cos(\sigma_i)
\end{aligned}
\end{equation}}{[R1-m4]} \par
The \textit{joint} vector space $\mb{q} = [q_1, q_2, q_3, q_4]$, representing the pulling/release displacement of the pulling wires, can be expressed using \eqref{eqn: L_i expre}, i.e., $q_i = L - L_i$. Then we define the \textit{joint-to-shape} space Jacobian $\mb{J}_{\mb{q}\mb{\mathcal{S}}}$ as the Jacobian relating infinitesimal $\delta \mb{q}$ to $\delta \mb{\mathcal{S}}$, as:
\begin{equation}
    \label{J_q_S}
    \mb{J}_{\mb{q}\mb{\mathcal{S}}} = \frac{\partial{\mb{q}}}{\partial{\mb{\mathcal{S}}}}, \qquad \mb{J}_{\mb{q}\mathcal{S}}\in \realfield^{4\times4}
\end{equation}
Jacobian $\mb{J}_{\mb{q}\mathcal{S}}$ is utilized to establish \textit{joint} space commands based on the control inputs in \textit{shape} state space during experiments.
The \editcomment{differential}{[R1-m10]} kinematics of \textit{joint-to-shape space} is given as
\begin{equation}
    \dot{\mb{q}}=\mb{J}_{\mb{q} \mb{\mathcal{S}}} \dot{\mb{\mathcal{S}}}
\end{equation}
By taking derivative of \eqref{eqn: L_i expre} with respect to the shape state $\mathcal{S}$, we thereby obtain the explicit expression for $\mb{J}_{\mb{q} \mb{\mathcal{S}}}$ as:
\begin{equation}
\mb{J}_{\mb{q} \mb{\mathcal{S}}}= r\left[\begin{array}{cccc}
 \cos (\sigma_1) & \frac{1}{2} \cos (\sigma_1) & \frac{1}{3} \cos (\sigma_1) & - \theta_e\sin (\sigma_1) \\ [2pt]
 \cos (\sigma_2) & \frac{1}{2} \cos (\sigma_2) & \frac{1}{3} \cos (\sigma_2) & - \theta_e\sin (\sigma_2) \\ [2pt]
 \cos (\sigma_3) & \frac{1}{2} \cos (\sigma_3) & \frac{1}{3} \cos (\sigma_3) & - \theta_e\sin (\sigma_3) \\ [2pt]
 \cos (\sigma_4) & \frac{1}{2} \cos (\sigma_4) & \frac{1}{3} \cos (\sigma_4) & - \theta_e\sin (\sigma_4) 
\end{array}\right]
\end{equation}
where $\theta_e = \theta_{(s=1)} = m_0 + \frac{1}{2}m_1 + \frac{1}{3}m_2$
\subsection{Mapping Between Shape and Task Spaces}
\label{sec:shape_Jac deriv}

\editcomment{
The task-space vector is defined as 
$\mb{x} \triangleq \big[\mb{p}\T,\; \bs{\omega}\T\big]\T \in \realfield^{6}$,
where $\mb{p}$ denotes the tip position expressed in frame $\{b\}$, and $\bs{\omega}$ denotes the axis--angle orientation coordinates of frame $\{g\}$ with respect to frame $\{b\}$.
}{[R1-m14]}

The \textit{shape-to-task space} Jacobian $\mb{J}_{\mb{x} \mathcal{S}}$ is formulated as
\begin{equation}
\mb{J}_{\mb{x} \mathcal{S}} = 
\left[\begin{array}{l}
\mb{J}_{\mb{p} \mathcal{S}} \\
\mb{J}_{\omega \mathcal{S}}
\end{array}\right], \qquad 
\mb{J}_{\mb{p}\mathcal{S}}\in\realfield^{3\times4}, \;\mb{J}_{\bs{\omega}\mathcal{S}}\in\realfield^{3\times4}
\end{equation}

\removecomment{
where Jacobian $\mb{J}_{\mb{p} \mathcal{S}}$ relates the small changes of position ${ }^b \mb{p}$ at the robot tip expressed in frame $b$  to the small changes in the shape state $\mathcal{S}$. Jacobian $\mb{J}_{\omega \mathcal{S}}$ relates small rotational changes (axis angle representation) of frame $g$ expressed w.r.t. frame $b$ to the small changes in the shape state $\mathcal{S}$.
}{}

\editcomment{
Here, $\mb{J}_{\mb{p}\mathcal{S}}$ relates infinitesimal changes in tip position to variations in the shape state $\mathcal{S}$, and $\mb{J}_{\omega\mathcal{S}}$ relates infinitesimal axis--angle orientation variations to changes in $\mathcal{S}$.
}{[R1-m14]}\par
Jacobian $\mb{J}_{\mb{p} \mathcal{S}}$ can be found by taking derivative of ${ }^b \mb{p}$ with respect to the shape state $\mathcal{S}$:
\editcomment{\begin{align}
&{ }^b \mb{p}={ }^b \mb{R}_1{ }^1 \mb{p} = \mb{R}_z(-\delta) ^1\mb{p}\\
&\mb{J}_{\mb{p} \mathcal{S}}=\frac{\partial ^b\mb{p}(1)}{\partial \mathcal{S}}=\left[\begin{array}{llll}
\frac{\partial ^b\mb{p}}{\partial m_0} & \frac{\partial ^b\mb{p}}{\partial m_1} & \frac{\partial ^b\mb{p}}{\partial m_2} & \frac{\partial ^b\mb{p}}{\partial \delta}
\end{array}\right]_{s=1} \\[4pt]
&\tfrac{\partial ^b\mb{p}(1)}{\partial m_i}=\mb{R}_{z(-\delta)}\left[L \int_0^1\left[\begin{array}{c}
\cos (\theta(s)) \\
0 \\
-\sin (\theta(s))
\end{array}\right] \frac{s^{i+1}}{i+1} \rmd s\right] \\[4pt]
&\tfrac{\partial ^b\mb{p}(1)}{\partial \delta}=\mb{R}_{z(-\delta)}\left[L \int_0^1\left[\begin{array}{c}
0 \\
-\sin (\theta(s)) \\
0
\end{array}\right] \rmd s\right] 
\end{align}}{[R1-m11]}
\editcomment{The above equations provide a semi-analytic expression for $\mb{J}_{\mb{p}\mathcal{S}}$: the dependence on the shape variables is explicit, while the remaining 1D integral terms are evaluated numerically and therefore do not require finite-difference differentiation.}{R1-m12}
To derive $\mb{J}_{\omega \mathcal{S}}$, we leverage the angular velocity of \editcomment{frame $\{g\}$}{[R1-m8]}, expressed as:
\begin{equation}
{ }^b \bs{\omega}_{g}=-\dot{\delta} \; \hat{\mb{z}}_b+{ }^b \mb{R}_1\left(\dot{\theta}(s) \;\mb{e}_2+{ }^1 \mb{R}_{es}\left(\dot{\delta} \;\mb{e}_3\right)\right)
\label{eqn_angular_v_theta_delta}
\end{equation}
\TopTextCushion
Linear relations can be expressed as $\dot{\delta}=\mb{A}\dot{\mathcal{S}}$ and $\dot{\theta}=\mb{B}\dot{\mathcal{S}}$. By rewriting \eqref{eqn_angular_v_theta_delta} in the form ${ }^b \bs{\omega}_{g}=\mb{J}_{\bs{\omega} \mathcal{S}}\dot{\mathcal{S}}$ and by applying the linear relations above, we extract the final expression as:

\begin{equation}
\mb{J}_{\bs{\omega} \mathcal{S}}=\left[\begin{array}{cccc}
-c_\theta s_\delta & -\frac{1}{2}c_\theta s_\delta & -\frac{1}{3}c_\theta s_\delta & s_\theta c_\delta \\ [2pt]
c_\theta c_\delta & \frac{1}{2}c_\theta c_\delta & \frac{1}{3}c_\theta c_\delta & s_\theta s_\delta \\ [2pt]
-s_\theta & -\frac{1}{2}s_\theta & -\frac{1}{3}s_\theta & c_\theta-1
\end{array}\right]
\end{equation}

\subsection{Solving for Modal Coefficients in Shape State}
Modal coefficients (e.g. $m_0$, $m_1$, $m_2$) must be solved in real-time to apply the virtual-work-based method effectively. We develop a solver leveraging the real-time inverse kinematics formulations to enhance shape estimation by minimizing the difference between observed and predicted pose at the tip. The objective function $Q(\mathcal{S})$ is defined as follows:
\begin{equation}
\label{eq:pose-error}
\min_{\mathcal{S}} Q(\mathcal{S})
=\tfrac{1}{2}
\left(\left\|
W
\begin{bmatrix}
\mb{p}_{\text{pred}}(\mathcal{S}) - \mb{p}_{\text{obs}} \\
\mathrm{Log}\!\big(\mb{R}_{\text{obs}}^{\top}\mb{R}_{\text{pred}}(\mathcal{S})\big)^{\vee}
\end{bmatrix}
\right\|_2\right)^2
\end{equation}
where 
\begin{equation}
    W=\mathrm{diag}(w_p,w_p,w_p,\,w_R,w_R,w_R)
\end{equation}
Here ${}^{b}\!\mb{T}_g(1)_{\text{obs}}$ and ${}^{b}\!\mb{T}_g(1)_{\text{pred}}(\mathcal{S})$ denote the observed and predicted tip transforms, with rotation components $\mb{R}_{\text{obs}},\ \mb{R}_{\text{pred}}$ and translation components $\mb{p}_{\text{obs}},\ \mb{p}_{\text{pred}}(\mathcal{S})$. 
The operator $\mathrm{Log}(\cdot)$ is the matrix logarithm mapping from $SO(3)$ to $\mathfrak{so}(3)$, and $(\cdot)^{\vee}$ maps a skew-symmetric matrix to its vector form in $\mathbb{R}^3$. 
The weights $w_p$ (m$^{-1}$) and $w_R$ (rad$^{-1}$) balance translation and rotation errors and provide unit consistency; for example, setting $w_R = 1/L$ with $L$ the instrument length makes a $1$ rad orientation error comparable to $L$ meters of position error. 
\removecomment{The optimization problem is solved using a quasi-Newton method that approximates the Hessian matrix of the objective function to achieve faster convergence.}\editcomment{The optimization problem is solved numerically as a nonlinear least-squares problem on the tip-pose residual. In the current implementation, the predicted tip pose is evaluated using a fast quadrature-based forward model, and the residual Jacobian with respect to the shape state is explicitly supplied using the semi-analytic Jacobian derived in Section~\ref{sec:shape_Jac deriv} together with the chain rule for the SO(3) geodesic rotation residual. In the current Python implementation, the Jacobian-enabled solver achieves an average solve time of approximately $3$\,ms for the final PCK2 stage---over $200\times$ faster than a finite-difference Jacobian baseline on the same hardware---making it well suited for real-time operation.}{R1-m13}\par
\editcomment{To improve robustness against local minima and parameter
non-uniqueness, the shape optimizer is initialized in a cascade
manner: the lower-order solution initializes the higher-order one
(PCK0 $\to$ PCK1 $\to$ PCK2), and in sequential quasi-static
operation the previous-step estimate serves as a warm start.
This continuation strategy provides a stable and practically
robust optimization for the present low-dimensional,
overdetermined setting.}{R2-C2} \par
\editcomment{Note that the shape estimation stage is a pure kinematics-based approach. It minimizes the discrepancy between the observed and predicted tip pose without invoking any static equilibrium or material properties. The physics-based virtual-work model enters exclusively in the second stage (Section \ref{sec:force-sensing model}), where the reconstructed shape state $\mathcal{S}$ is used in the force balance.}{[R1-m2]}
\subsection{Virtual-work Based Force Sensing Model}
\label{sec:force-sensing model}
The elastic bending strain energy density at location $s$ can be expressed as:
\begin{equation}
u(s) = \frac{EI}{2L} \kappa(s)^2 = \frac{EI}{2L} \left(m_{0} + m_{1}s + m_{2}s^2\right)^2
\end{equation}
where \( E \) and \( I \) are the Young's modulus and the second moment of area of the central backbone, respectively. 
The elastic energy $U$ stored in the continuum instrument, with normalized arclength \( s \in(0,1)\), is obtained via an integral:
\begin{equation}
U = \frac{EI}{2L} \int_0^1 \kappa(s)^2 ds
\label{eqn: elastic_energy}
\end{equation} \par
Inspired by \cite{xu2008investigation,zhao2022modular}, we applied the virtual work principle to build a static equilibrium equation with respect to the shape state $\mathcal{S}$ as:
\begin{equation}
\label{eqn:static equilibrium}
(\mb{J}_{\mb{q} \mathcal{S}})\T \;\bs{\tau}\; +\; (\mb{J}_{\mb{x} \mathcal{S}})\T \;\mathcal{F}\; =\; \nabla_{\mathcal{S}} U
\end{equation}
where $\mathcal{F}$ represents the external wrench acting at the tip of the continuum robot, while $\bs{\tau}$ denotes the actuation forces (i.e., cable tensions). In practice, cable tensions are bounded by mechanical limits
that can be determined from the present virtual-work formulation
for a given bending configuration and external load. $\nabla_{\mathcal{S}} U$ is the gradient of the elastic energy, obtained by differentiating \eqref{eqn: elastic_energy} with respect to $\mathcal{S}$:
\begin{equation}
\nabla_{\mathcal{S}} U \; \triangleq \;
\frac{\partial U}{\partial \mathcal{S}} \; = \; 
\frac{EI}{L} \begin{bmatrix}
m_{0} + \tfrac{1}{2} m_{1} + \tfrac{1}{3} m_{2} \\[2pt]
\tfrac{1}{2} m_{0} + \tfrac{1}{3} m_{1} + \tfrac{1}{4} m_{2} \\[2pt]
\tfrac{1}{3} m_{0} + \tfrac{1}{4} m_{1} + \tfrac{1}{5} m_{2} \\[2pt]
0
\end{bmatrix}
\label{eqn_grad_U_shape}
\end{equation}
This gradient represents the internal elastic forces resisting deformation, providing a key link between shape states and external forces acting on the continuum robot.\par 

The static equilibrium governing equation \eqref{eqn:static equilibrium} represents a redundancy resolution problem for determining the external wrench \(\mathcal{F}\) that satisfies it. Its general solution is given by:
\begin{equation}
\hspace{2mm}
\mathcal{F}
=
\underbrace{\bigl(\mb{J}_{\mb{x}\mathcal{S}}\T\bigr)^{+}
\bigl(\nabla_{\mathcal{S}}U-\mb{J}_{\mb{q}\mathcal{S}}\T\boldsymbol{\tau}\bigr)}_{\displaystyle\mathcal{F}_{\mathrm{p}}}
+
\underbrace{\Bigl(\mb{I}-\bigl(\mb{J}_{\mb{x}\mathcal{S}}\T\bigr)^{+}
                \mb{J}_{\mb{x}\mathcal{S}}\T\Bigr)}_{\displaystyle\mb{N}}
\boldsymbol{\eta}
\label{eq:general_sol}
\end{equation}
where  
\(\mathcal{F}_{\mathrm{p}}\) is the particular solution and 
\(\mb{N}\bs{\eta}\) is the homogeneous solution involving the null space projector \(\mb{N}\).
The redundancy is resolved by an arbitrary selection of \(\boldsymbol{\eta}\).\par
\paragraph*{Weighted least-squares criterion}
To select a unique external wrench, we minimize the weighted deviation from a prior estimate \(\mathcal{F}_{0}\) using a weighting matrix
\(\mb{W}\succcurlyeq0\):
\begin{equation}
\boldsymbol{\eta}^{*}
=
\arg\min_{\boldsymbol{\eta}}
\bigl(\mathcal{F}_{\mathrm{p}}+\mb{N}\boldsymbol{\eta}-\mathcal{F}_{0}\bigr)\T
\mb{W}\,
\bigl(\mathcal{F}_{\mathrm{p}}+\mb{N}\boldsymbol{\eta}-\mathcal{F}_{0}\bigr)
\label{eq:weighted_LS}
\end{equation}
\paragraph*{Closed-form solution}
Solving the quadratic optimization problem in \eqref{eq:weighted_LS} yields the optimized solution \(\bs{\eta}^*\) using the Moore–Penrose pseudoinverse:
\begin{equation}
\boldsymbol{\eta}^{*}
=
\mb{\Omega}^{+}\,
\mb{N}\T\mb{W}\,
\bigl(\mathcal{F}_{0}-\mathcal{F}_{\mathrm{p}}\bigr), \quad\mb{\Omega}=\mb{N}\T\mb{W}\mb{N}
\label{eq:eta_closed}
\end{equation}
Substituting \(\boldsymbol{\eta}^{*}\) back into \eqref{eq:general_sol} provides the unique wrench solution consistent with the prior estimate:
\begin{equation}
\mathcal{F}_{*}
=
\mathcal{F}_{\mathrm{p}}
+
\mb{N}\,
\mb{\Omega}^{+}\,
\mb{N}\T\mb{W}\,
\bigl(\mathcal{F}_{0}-\mathcal{F}_{\mathrm{p}}\bigr)
\label{eq:final_wrench}
\end{equation}

% \smallskip
\editcomment{The present force-sensing formulation is defined within the deformation subspace described by $\mathcal{S}=[m_0,m_1,m_2,\delta]^\top$. Although the tip wrench is written in 6D task-space form, only the components compatible with the retained planar bending coordinates are meaningfully identified under the available measurements; extension to torsional and general out-of-plane deformation would require a higher-dimensional backbone model with richer sensing or additional constraints.}{R1-M2,R1-M3,R2-C2}
\removecomment{The redundancy resolution in \mbox{\eqref{eq:final_wrench}}{} follows the same weighted least-squares strategy as the seminal work of
\mbox{\cite{xu2008investigation}}{}, but differs in the underlying shape
parameterization: We express the shape directly in curvature space rather than through the constant-curvature configuration variables \mbox{\((\theta,\delta)\)}{} employed in \mbox{\cite{xu2008investigation}}{}.  This relaxes the circular-arc assumption and accommodates general bending profiles.}{}
\editcomment{The redundancy resolution in \eqref{eq:final_wrench} follows the same weighted least-squares strategy as \cite{xu2008investigation}, but expressed directly in curvature space rather than through the constant-curvature variables $(\theta,\delta)$, thereby relaxing the circular-arc assumption. Because $\mb{J}_{\mb{x} \mathcal{S}}\in\mathbb{R}^{6\times 4}$ is rank-deficient, this step should be interpreted as regularization-based redundancy resolution rather than as rendering null-space wrench components observable. The prior selects a unique solution consistent with the assumed planar operating regime, not to infer axial torsion or other unobservable components from measurements alone.}{R2-C2,R1-m15}

We adopt a second-order polynomial curvature representation (PCK2):$\kappa(s)=m_0 + m_1 s + m_2 s^{2},$
and use it consistently in both the forward kinematics and the
virtual-work Jacobians \(\mb{J}_{\mb{q}\mathcal{S}}\) and \(\mb{J}_{\mb{x}\mathcal{S}}\).  Reducing the polynomial order to zero recovers the constant curvature
model (PCK0), and the first order case (PCK1) allows monotonic curvature variation along the backbone.

The second order form (PCK2) is selected as the representation with higher fidelity, capable of capturing non-uniform or partially inflected bending profiles while still admitting \editcomment{semi-analytic}{R1-m12} Jacobians and an analytic redundancy resolution.

\section{Simulation Design}
\begin{figure*}[t!]
    \vspace*{1.0ex}
    \centering
    \includegraphics[width=1\columnwidth]{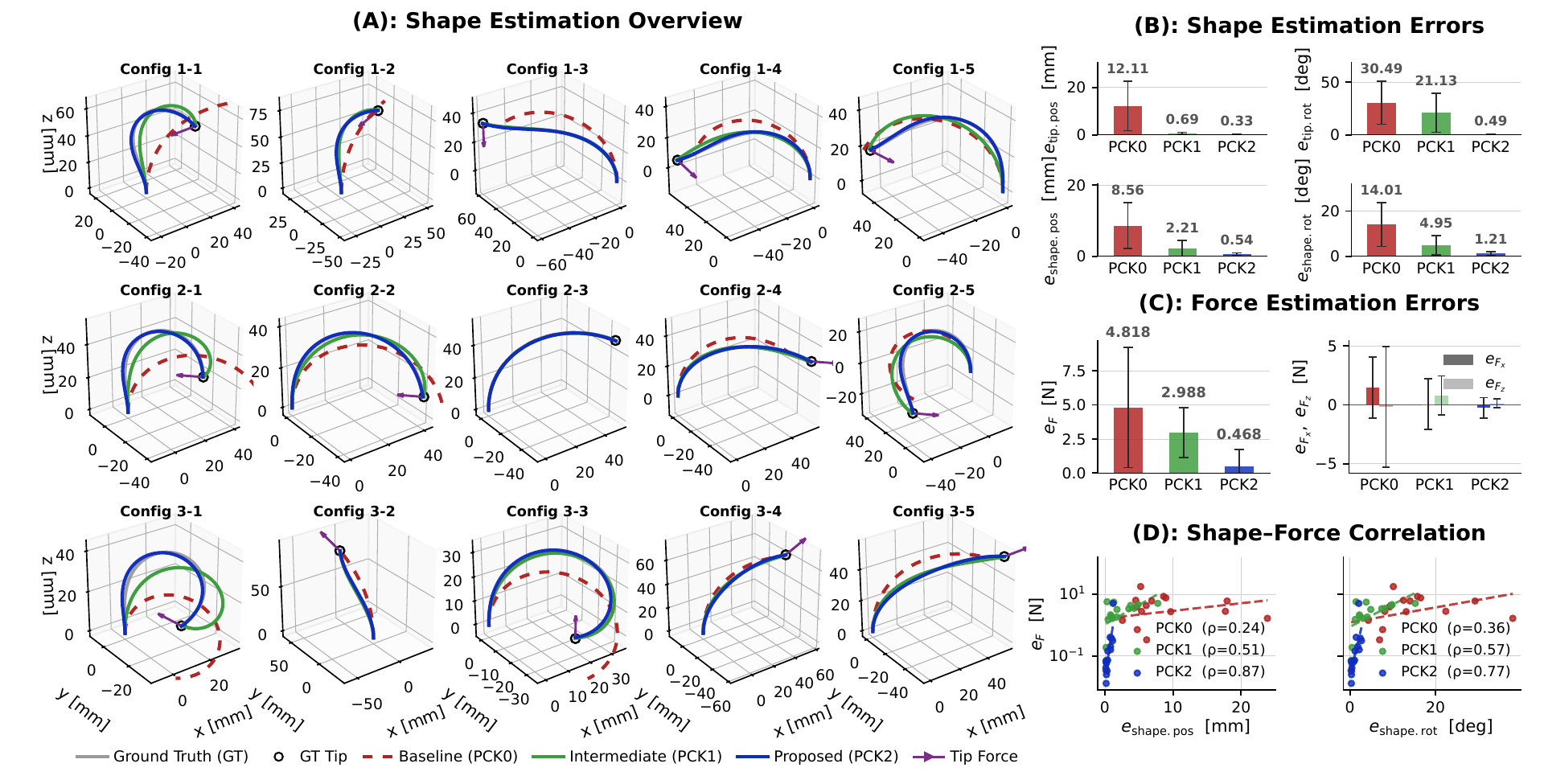}
    \caption{\editcomment{Shape and force estimation performance across fifteen external loading configurations arranged in a $3\times5$ grid (rows: $F_z \in \{-2, 0, +2\}$~N; columns: $F_x \in \{-4,-2,0,+2,+4\}$~N) under actuation tensions $\tau=[5,0,0,5]$~N with measurement noise. (A) Shape estimation overview. The ground-truth backbone is shown by a thin gray line with a black marker at the tip. PCK0 (baseline), PCK1, and the proposed PCK2 reconstructions are overlaid as red dashed, green, and blue curves, respectively. A purple arrow at each tip indicates the applied force vector. (B) Aggregate shape estimation errors (mean $\pm$ std over configurations), including tip and shape position errors as well as orientation errors. (C) Force estimation errors summarized by the force magnitude error $e_F=\|\hat{\mathbf{F}}-\mathbf{F}\|_2$ and component-wise errors. (D) Correlation between shape and force estimation errors across configurations. Both position-based $e_{\text{shape,pos}}$ and orientation-based $e_{\text{shape,rot}}$ shape errors exhibit clear positive correlation with force error. The proposed PCK2 method shows the strongest and most consistent coupling (highest $\rho$), indicating that improved shape reconstruction directly translates into more accurate force estimation.}{R1-M3}}
    \label{fig:sim-est-results}
\end{figure*}
This section evaluates the proposed estimator in a controlled simulation environment before physical validation.
The goal is to assess shape and force estimation accuracy under known boundary conditions, varying loading, and different curvature profiles.\par
\subsection{Settings}
A planar Cosserat rod solver was used to generate fifteen ground truth backbone shapes under known combinations of external tip force and cable tensions. Each case modeled a single segment with length \(L=100\,\mathrm{mm}\), elastic modulus \(E=6\times10^{10}\,\mathrm{Pa}\), and a circular cross section of radius \(r=0.5\,\mathrm{mm}\). Four actuation cables were evenly routed around the backbone at a radial offset \(R_p=8\,\mathrm{mm}\).\par
\removecomment{The fifteen configurations corresponded to distinct tip-force conditions defined by the Cartesian components \mbox{\(F_x\in\{-2,-1,0,1,2\}\,\mathrm{N}\)} and \mbox{\(F_z\in\{+1,0,-1\}\,\mathrm{N}\)}.}{}
\editcomment{The fifteen configurations corresponded to planar tip-force conditions defined in the bending-plane frame $\{1\}$, with components \(F_{x_1}\in\{-4,-2,0,+2,+4\}\,\mathrm{N}\) and \(F_{z_1}\in\{-2,0,+2\}\,\mathrm{N}\).}{R1-M3}
\removecomment{In all cases, the actuation tensions were fixed at \mbox{\(\boldsymbol{\tau}=[5,\,5,\,0,\,0]\,\mathrm{N}\)} with a bending plane heading of \mbox{\(\delta=-45^\circ\)}}{}\editcomment{In all cases, the actuation tensions were fixed at \(\boldsymbol{\tau}=[5,\,0,\,0,\,5]\,\mathrm{N}\) with a bending plane heading of \(\delta=45^\circ\)}{}. Synthetic measurements were formed by perturbing the true tip pose with zero mean Gaussian noise \((\sigma_x=\sigma_z=0.5\,\mathrm{mm},\ \sigma_\theta=0.005\,\mathrm{rad})\). The tensions used in the virtual work force solve were likewise corrupted with zero mean Gaussian noise \((\sigma_\tau=0.02\,\mathrm{N})\) and a small bias drift per trial, and each \(\tau_i\) was clipped to satisfy \(\tau_i\ge 0\).\par
The synthetic measurements, comprising tip pose and cable tensions, were then input to the estimation framework to reconstruct the backbone shape and tip force. Polynomial curvature (PCK) models of orders 0 through 2 were evaluated. For each configuration, five independent trials were conducted. Performance is quantified using the following metrics: tip errors \(e_{\text {tip,pos}}=\|\hat{\boldsymbol{p}}(1)-\boldsymbol{p}(1)\|_2\) and \removecomment{\mbox{\(e_{\text{tip,rot}}=|\hat{\theta}(1)-\theta(1)|\)}}{}\editcomment{$e_{\text {tip,rot }}=\cos ^{-1}\left(\frac{\operatorname{tr}\left(\left({ }^b R_g(1)\right)^{\top}{ }^b \hat{R}_g(1)\right)-1}{2}\right)$}{}; shape (arclength-averaged) errors with \(N\) uniformly sampled points \(s_i\in[0,1]\), \(e_{\text {shape,pos}}=\tfrac{1}{N}\sum_{i=1}^{N}\|\hat{\boldsymbol{p}}(s_i)-\boldsymbol{p}(s_i)\|_2\) and \removecomment{\mbox{\(e_{\text{shape,rot}}=\tfrac{1}{N}\sum_{i=1}^{N}|\hat{\theta}(s_i)-\theta(s_i)|\)}}{}\editcomment{$e_{\text {shape,rot }}=\frac{1}{N} \sum_{i=1}^N \cos ^{-1}\left(\frac{\operatorname{tr}\left(\left({ }^b R_g\left(s_i\right)\right)^{\top}{ }^b \hat{R}_g\left(s_i\right)\right)-1}{2}\right)$}{}; and force errors \(\boldsymbol{e}_F=\hat{\boldsymbol{F}}-\boldsymbol{F}\), with scalar magnitude \(e_F=\|\boldsymbol{e}_F\|_2\) and components \(e_{F_x}=\hat{F}_x-F_x\), \(e_{F_z}=\hat{F}_z-F_z\).\par

\subsection{Results}

\removecomment{Qualitative shape reconstructions are shown in \mbox{Fig.~\ref{fig:sim-shape-est-results}} in a \mbox{\(3\times5\)} grid layout. The absolute mean error (MAE) of shape estimation is summarized in \mbox{Table~\ref{tab:shape_overall_mae}}. Force estimation errors for each configuration are reported in \mbox{Table~\ref{tab:force_per_config_compact}}, and the overall MAE is summarized in \mbox{Table~\ref{tab:force_overall_mae}}.}{}\editcomment{The simulation results are consolidated in \mbox{Fig.~\ref{fig:sim-est-results}}, which summarizes qualitative shape reconstruction, aggregate shape-estimation errors, force-estimation errors, and shape--force error correlation across all fifteen loading configurations.}{R1-M3}

\removecomment{As shown in \mbox{Fig.~\ref{fig:sim-shape-est-results}}, all estimators qualitatively reconstructed the backbone geometry across the fifteen external-force configurations. The PCK2 model exhibited the closest agreement with the ground truth backbone, maintaining curvature continuity even under large bending and mixed loading. In contrast, PCK0 systematically underestimated curvature and failed to capture the nonlinear deformation when lateral and axial forces were applied. PCK1 alleviates most of this limitation and performs comparably to PCK2 across the majority of configurations, but noticeable shape deviations appear in Configs. 2-1 and 3-1.}{}
% \editcomment{As shown in \mbox{Fig.~\ref{fig:sim-est-results}(A)}, all three estimators recover the overall backbone deformation trend across the fifteen loading configurations. The proposed PCK2 model provides the closest agreement with the ground-truth shape throughout the loading set, while PCK0 tends to underfit the curvature and exhibits the largest geometric deviations under mixed loading. PCK1 improves the reconstruction substantially and remains close to PCK2 in many cases, although visible discrepancies still appear in several more demanding configurations.}{R1-M3}

\removecomment{Quantitatively, \mbox{Table~\ref{tab:shape_overall_mae}} reports the mean absolute errors (MAE) of shape estimation across all configurations. Increasing the polynomial order yielded consistent improvement in both position and rotation accuracy. The arclength-averaged position error \mbox{\(e_{\text{shape,pos}}\)} decreased from \mbox{\(8.12\,\mathrm{mm}\)} (PCK0) to \mbox{\(0.20\,\mathrm{mm}\)} (PCK2), and the rotation error \mbox{\(e_{\text{shape,rot}}\)} decreased from \mbox{\(9.06^{\circ}\)} to \mbox{\(0.38^{\circ}\)}. Similar trends were observed for tip errors, where \mbox{\(e_{\text{tip,pos}}\)} dropped from \mbox{\(13.87\,\mathrm{mm}\)} to \mbox{\(0.21\,\mathrm{mm}\)}. These results confirm that higher-order curvature representations significantly enhance shape reconstruction fidelity.}{}
% \editcomment{The aggregate statistics in \mbox{Fig.~\ref{fig:sim-est-results}(B)} show that increasing the polynomial order consistently reduces both position and rotation errors. In particular, the mean tip-position error decreases from 12.11 mm (PCK0) to 0.69 mm (PCK1) and 0.33 mm (PCK2), while the mean tip-rotation error decreases from 30.49$^\circ$ to 21.13$^\circ$ and 0.49$^\circ$, respectively. A similar trend is observed for the arclength-averaged shape errors: \(e_{\text{shape,pos}}\) decreases from 8.56 mm to 2.21 mm and 0.54 mm, and \(e_{\text{shape,rot}}\) decreases from 14.01$^\circ$ to 4.95$^\circ$ and 1.21$^\circ$. These results confirm that higher-order curvature representations significantly improve shape reconstruction fidelity.}{R1-M3}

\removecomment{Force estimation performance, summarized in \mbox{Tables~\ref{tab:force_per_config_compact} and~\ref{tab:force_overall_mae}}, shows a consistent correlation between shape accuracy and recovered tip forces. Across all fifteen cases, the PCK2 model achieved the lowest overall force MAE of \mbox{\(0.05\,\mathrm{N}\)}, compared to \mbox{\(0.12\,\mathrm{N}\)} for PCK1 and \mbox{\(8.22\,\mathrm{N}\)} for PCK0. Component-wise errors averaged \mbox{\(0.03\,\mathrm{N}\)} along the x and z directions. The per-configuration results indicate stable performance across different external load directions, with no apparent bias to the force orientation. Together, these findings demonstrate that the proposed estimation framework reliably reconstructs both shape and external force from noisy measurements.}{}
\removecomment{We examined how shape and force errors co-vary using the Pearson correlation coefficient \mbox{\(\rho\)}, which ranges from \mbox{\(-1\)} to \mbox{\(+1\)} (\mbox{\(\rho\approx 0\)} indicates no clear linear trend). As shown in \mbox{Fig.~\ref{fig:shape-force-err-correlation}}, PCK2 achieves uniformly small shape and tip errors, resulting in weak apparent correlations despite its highest overall accuracy. For PCK0, shape and force errors increase consistently together, indicating that geometric deviations directly amplify force-estimation errors. PCK1 exhibits intermediate behavior---moderate correlations among shape and tip metrics. These results indicate that PCK2 attains the most accurate shape and force estimation among all models. The remaining variation in its force estimates suggests that force accuracy is not solely determined by geometric reconstruction but is also influenced by factors such as measurement noise, parameter uncertainty, and unmodeled actuation effects.}{}
\editcomment{As shown in Fig.~\ref{fig:sim-est-results}(A), all three estimators recover the overall backbone deformation trend.
The proposed PCK2 model provides the closest agreement with the ground-truth shape throughout,
while PCK0 tends to underfit curvature under mixed loading and PCK1 remains intermediate
with visible discrepancies in more demanding configurations.
The aggregate statistics in Fig.~\ref{fig:sim-est-results}(B) confirm that increasing the polynomial order
consistently reduces both position and rotation errors:
the mean tip-position error decreases from $12.11$\,mm (PCK0) to $0.33$\,mm (PCK2),
and the arclength-averaged shape-position error from $8.56$\,mm to $0.54$\,mm.
Tip-rotation errors show a similar trend, dropping from $30.49^\circ$ to $0.49^\circ$.}{}

\editcomment{The force-estimation summary in Fig.~\ref{fig:sim-est-results}(C) follows the same ordering:
the mean force-magnitude error $e_F$ decreases from $4.818$\,N (PCK0)
to $2.988$\,N (PCK1) and $0.468$\,N (PCK2),
indicating that improved shape reconstruction directly benefits
virtual-work-based force recovery.
Figure~\ref{fig:sim-est-results}(D) further shows that PCK2 exhibits the strongest
Pearson correlation between shape and force errors
($\rho = 0.87$ for position, $\rho = 0.77$ for orientation),
supporting the central premise of the framework:
within the modeled planar loading regime,
more accurate curvature-space shape estimation translates directly
into more accurate force estimation.}{}

\section{Experimental Validation}
\label{sec:exp_validation}
\begin{figure}[t]  % no "!"
  % tiny cushion inside the text block (not a margin hack)
  \vspace*{1.0ex}
  \centering
  \includegraphics[
    width=0.75\linewidth,
    height=0.38\textheight,      % reduce if still tight: 0.36–0.37
    keepaspectratio,
    clip,trim=6pt 6pt 6pt 6pt    % trims white borders so the BOX is smaller
  ]{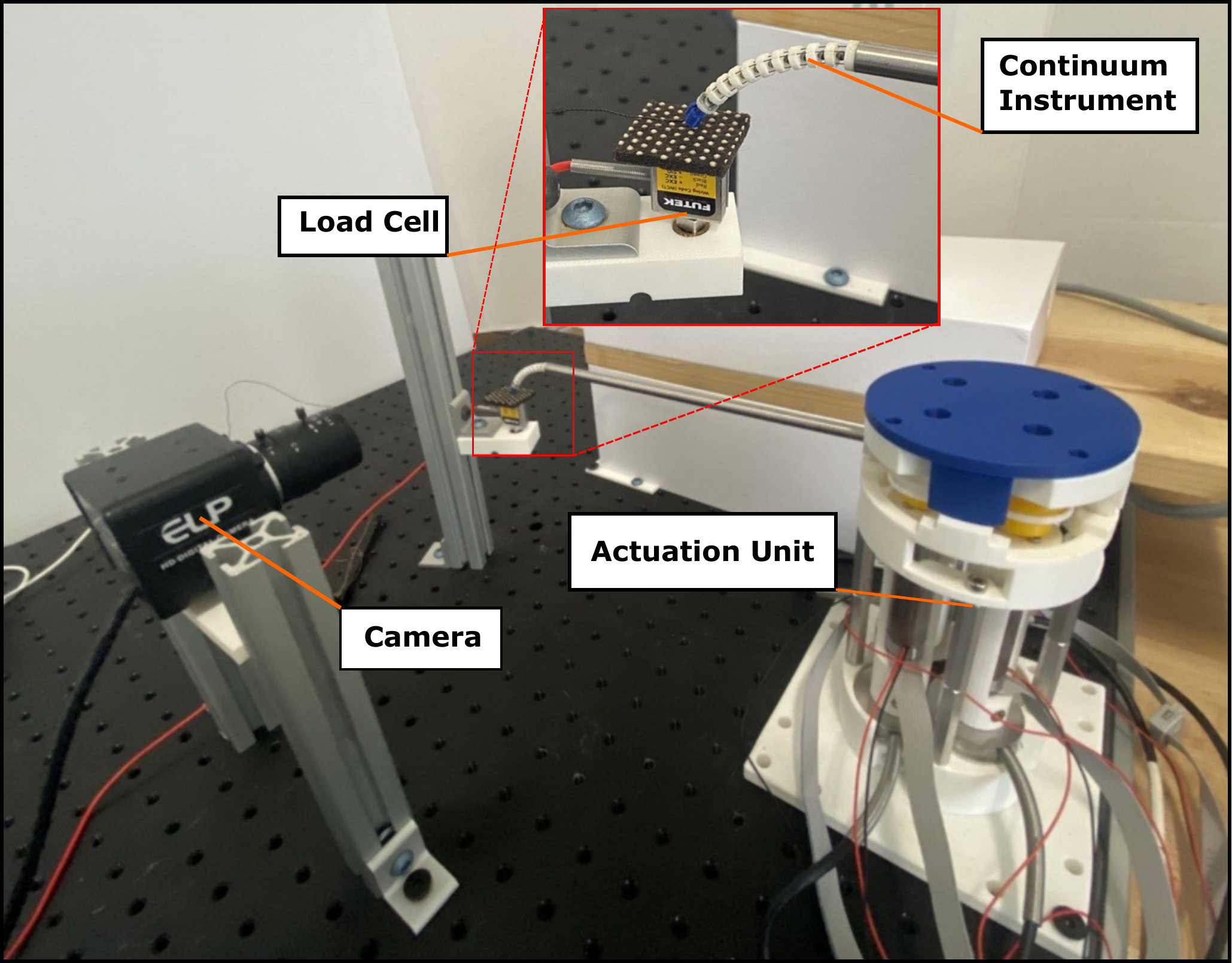}
  \caption{Experimental setup for validating shape and tip-force estimation. An ELP HD USB camera (1024\(\times\)768 px, 30 Hz) tracks the tip pose, while a miniature load cell (\(\pm{}1\;\text{N}\) range) records contact forces. The continuum instrument tested here has backbone length \(L = 40\;\text{mm}\), distance between central backbone and pulling wires \(r = 1.8\;\text{mm}\), Young’s modulus \(E = 65\;\text{GPa}\), and second moment of area \(I = 4.83 \times 10^{-15}\;\text{m}^4\).}
  \label{fig:shape-force-test-setup}
\end{figure}
\subsection{Experimental Setup}
This section validates our integrated shape–force estimation approach on a real continuum robot.
The experimental setup is shown in Fig.~\mbox{\ref{fig:shape-force-test-setup}}{}.
The platform consists of a customized continuum instrument and a torque-cell-enabled actuation unit.
The instrument is driven by four cables anchored at the distal tip, enabling planar and spatial bending similar to the simulation configuration. Each cable is actuated by a DC motor equipped with a small torque cell mounted behind the motor to measure the reaction torque on the 3D-printed bracket, from which cable tension is computed.
Four motor–torque modules are arranged in a cylindrical layout, providing modularity and compactness.
All motors are driven by a Speedgoat real-time target via EtherCAT through Elmo drives, ensuring synchronized acquisition of motor position and torque data for shape–force estimation experiments.
A USB camera (ELP Inc., 1024×768 px, 30 Hz) was used to capture the instrument’s backbone shape and tip pose. \editcomment{The USB camera is used here only as a benchtop validation tool in a controlled laboratory environment and is not intended as the sensing modality for surgical deployment.}{R2-C1}
Camera parameters were calibrated using a standard checkerboard under a pinhole model.
During experiments, the instrument was bent into multiple static configurations, and the backbone was manually segmented from images to obtain ground-truth shape data for validation.
Before attaching the continuum instrument, each motor–torque module was individually calibrated by hanging known weights from the pulley. \removecomment{This pre-assembly calibration compensates for friction and structural biases introduced by the motor bracket alignment, pulley installation, and cable routing through the rigid tube that houses the guide tunnels, ensuring accurate mapping from measured reaction torque to proximal cable tension.}{} \editcomment{This pre-assembly calibration partially compensates for frictional and structural biases introduced by motor-bracket alignment, pulley installation, and cable routing through the rigid tube that houses the guide tunnels.
The remaining mismatch---attributed mainly to distributed friction
and hysteresis between pulling wires and backbone disks during
bending---is not explicitly modeled in the present virtual-work
formulation and constitutes the primary residual source of
force-estimation error.}{R2-C3}
\subsection{Test Scenarios}
\label{sec:test_scenarios}

Three scenarios were designed to evaluate the proposed shape and force estimation methods.
Scenarios~I and~II were conducted under free bending.
Scenario~I uses \(\theta=67^\circ,\ \delta=135^\circ\), and Scenario~II uses \(\theta=45^\circ,\ \delta=-45^\circ\), where \(\theta\) is the bending angle and \(\delta\) is the orientation of the bending plane.
In our setup (Fig.~\mbox{\ref{fig:shape-force-test-setup}}{}), \(\delta=-45^\circ\) produces a downward bend in the camera view, whereas \(\delta=135^\circ\) produces an upward bend.

\removecomment{Scenario III keeps a similar configuration to Scenario II \mbox{\((\theta = 47.8^\circ, \delta = -45^\circ)\)} but introduces tip contact against a rubber sheet mounted on a miniature load cell \mbox{(see Fig.~\mbox{\ref{fig:shape-force-test-setup}}{})}.}{}\editcomment{Scenario III keeps a similar configuration to Scenario II \((\theta = 47.8^\circ, \delta = -45^\circ)\) but introduces controlled single-point tip contact against a rubber sheet mounted on a miniature load cell (see Fig.~4), representing a simple palpation-like contact case.}{R2-C3}
In all scenarios, two cables were pulled while the other two were released by equal displacements to generate the desired bending profiles.

For force estimation, four distinct cable-tension loading cases were tested under Scenario~III, yielding the same geometric configuration with different combinations of proximal tensions and external contact forces.
Each scenario was repeated ten times for both constant-curvature (CC/PCK0) and polynomial-curvature (PCK1 and PCK2) models to quantify accuracy and repeatability. \editcomment{In all trials, the robot was moved to the target configuration and held at rest during data acquisition, so the reported results correspond to static or near-static conditions.}{R2-C4}
\subsection{Results}
\label{sec:results}
\begin{figure}[!t]
    \centering
    \includegraphics[width=1\columnwidth]{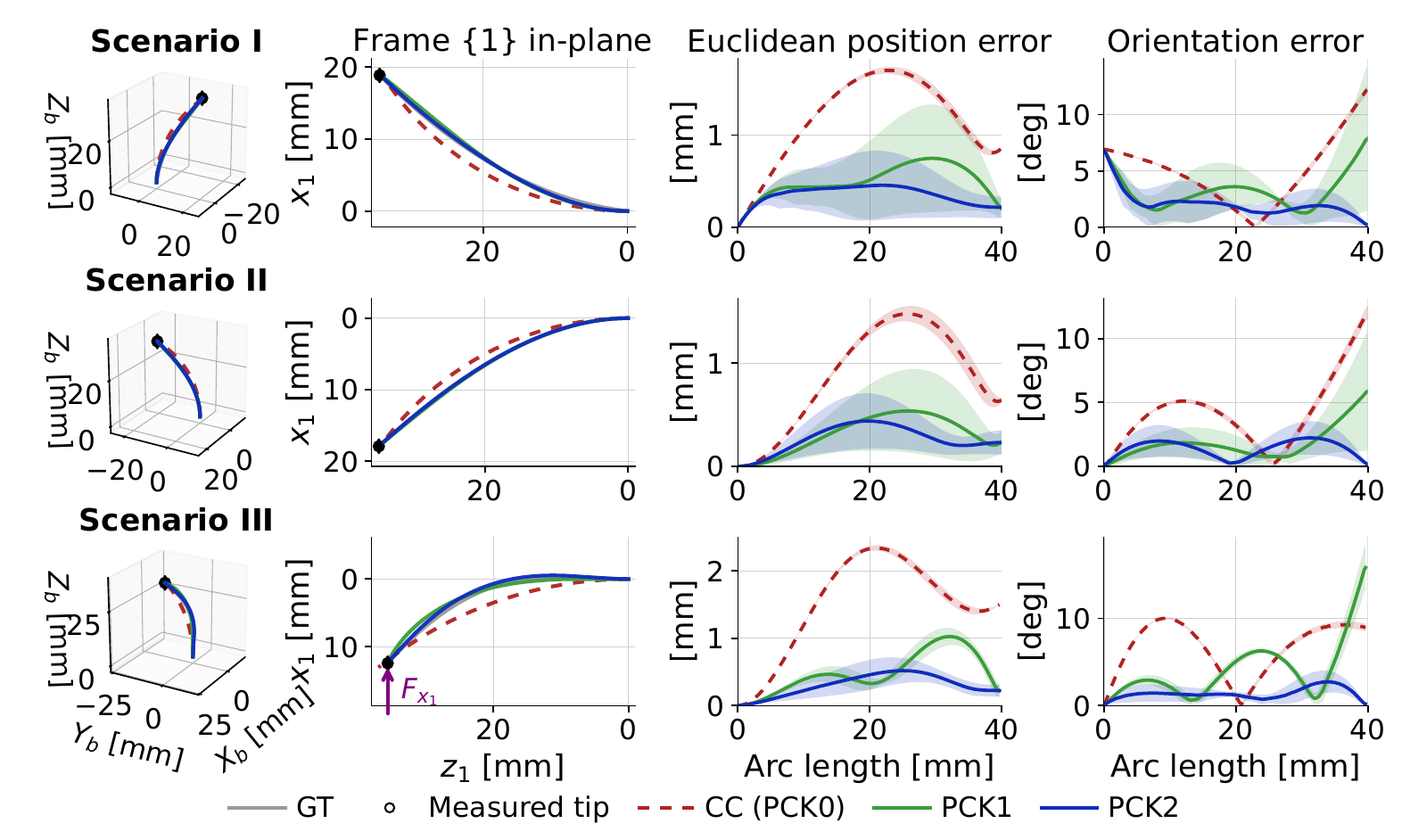}
    \caption{\editcomment{Experimental shape-estimation results under three scenarios. Each row corresponds to one scenario. The first two columns show the mean reconstructed backbone shape over ten trials, compared with the ground truth (GT) and the measured tip pose, expressed in frames $\{b\}$ and $\{1\}$, respectively. The third and fourth columns show the Euclidean position error and orientation error along the arclength, respectively (mean $\pm$ 1 std). Scenarios I and II are free-bending cases with different bending-plane orientations, while Scenario III corresponds to a controlled single-point tip-contact case against a rubber sheet.}{R2-m1}}
  \label{fig:shape_estimation_results}
\end{figure}
\begin{figure}[!h]
	\centering
	\includegraphics[width=1\columnwidth]{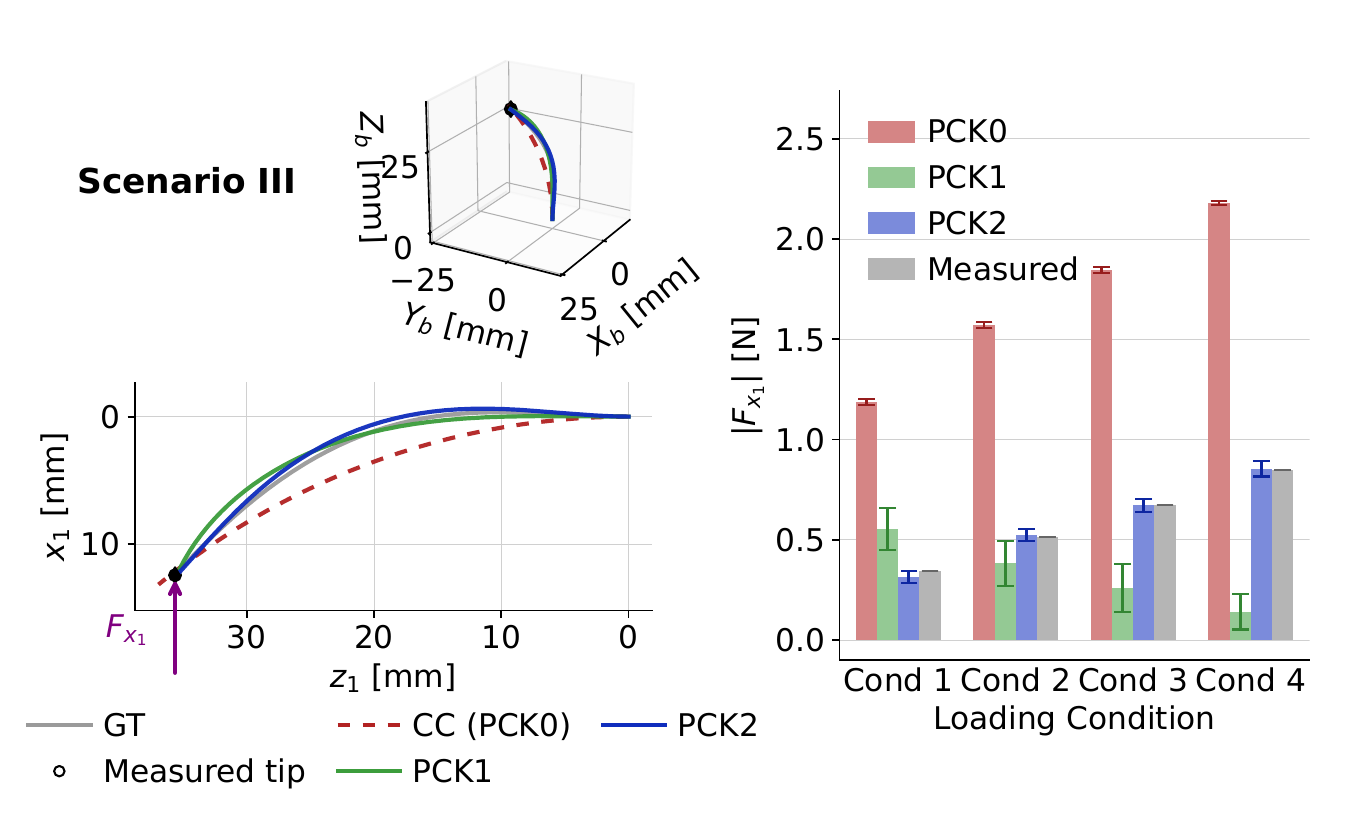}
    \caption{\editcomment{Experimental force-estimation results under Scenario III for four loading conditions (Cond.~1--Cond.~4). The left panel shows the mean reconstructed shapes in the contact configuration compared with the ground truth (GT) and measured tip pose. The right panel reports the mean magnitude of the bending-plane tip-force component $|F_{x_1}|$ over ten trials for each model, together with the load-cell measurement. Here, $F_{x_1}$ denotes the force component along the $x_1$-axis of frame $\{1\}$. Error bars indicate $\pm 1$ standard deviation.}{R1-M3,R1-m16}}
	\label{fig:force_estimation_results}
\end{figure}

Figure~\mbox{\ref{fig:shape_estimation_results}}{} summarizes the shape estimation performance.
\removecomment{In each scenario, the top row shows the mean reconstructed shapes compared with the ground truth and measured tip poses, while the middle and bottom rows show the Euclidean position and orientation errors along the arclength (\emph{mean}~$\pm$~1~std).}{}\editcomment{Each row corresponds to one scenario. The first and second columns show the mean reconstructed shapes compared with the ground truth and measured tip poses, while the third and fourth columns show the Euclidean position and orientation errors along the arclength (mean $\pm$ 1 std), respectively.}{} Across all scenarios, both PCK1 and PCK2 produce noticeably lower shape and orientation errors than the constant curvature model, with PCK2 providing the closest agreement to the ground truth. \removecomment{Small curvature variations observed even under free bending are attributed to practical factors such as tendon friction, routing misalignment, and assembly tolerance. Scenario III further confirms that the polynomial models maintain robustness under tip contact, where local curvature changes become more pronounced.}{}\editcomment{Small curvature variations observed even under free bending are attributed to practical factors such as tendon friction, routing misalignment, and assembly tolerance. Scenario III further confirms that the polynomial models maintain robustness under a simple single-point contact case, where local curvature changes become more pronounced.}{R2-C3}

\editcomment{In the present experiments, we report the bending-plane tip-force component $F_{x_1}$, i.e., the force component along the $x_1$-axis of frame $\{1\}$, because the loading and deformation remain within the modeled planar regime.}{R1-M3, R1-m16} Figure~\mbox{\ref{fig:force_estimation_results}}{} reports the corresponding tip-force estimates under Scenario III for four loading conditions (Cond1–Cond4).
Among the three models, PCK2 tracks the measured force most closely, while  PCK0/CC and PCK1 exhibit larger deviations.

Force estimation in the virtual-work formulation is highly
sensitive to distal-end shape accuracy, where small orientation
deviations amplify wrench errors.
Both PCK0 and PCK1 produce noticeable tip-orientation
bias---the latter because its monotonic curvature constraint
cannot capture deformation governed jointly by actuation and
external loading---distorting the strain-energy gradient and
causing systematic force estimation errors.

Consistent with simulation findings, the experimental results
confirm a strong coupling between shape and force accuracy.
The PCK2 model alleviates these effects by better accommodating
curvature variation, thereby reducing orientation bias and
improving the stability of the virtual-work mapping.
Overall, PCK2 provides the most accurate and consistent
estimation across all test scenarios.

\section{Conclusion}
This letter presented an integrated shape–force estimation framework for cable-driven continuum robots based on a virtual-work formulation expressed directly in curvature space. By combining tip-pose sensing with cable-tension measurements, the framework reconstructs backbone shape and estimates external tip forces through a two-stage pipeline under sparse sensing conditions.
The method employs polynomial curvature kinematics (PCK) to parameterize the backbone with a small set of modal coefficients, enabling semi-analytic Jacobians and closed-form strain-energy gradients. Simulation and experimental results consistently show that the second-order model (PCK2) achieves the highest accuracy in both shape and force estimation, confirming that curvature-space modeling yields a compact and robust alternative to geometry-space and constant-curvature approaches.
\editcomment{In simulation, PCK2 reduces the mean force-estimation error to $0.468$\,N (versus $4.818$\,N for constant curvature), while in experiment it tracks the load-cell measurement within the sensor noise floor across all tested conditions.}{}
\editcomment{The present formulation assumes static or near-static operation; future work will extend it to full spatial $(SE(3))$ configurations and faster dynamic motion, which will require a higher-dimensional backbone model together with richer sensing or additional constraints, as well as integration with real-time estimation and control frameworks for continuum robotic manipulation.}{R1-M2,R2-C2,R2-C4}

% \appendix
% \section{Open Source Code}
% \label{app:code}
% Open Source Code: The Python code developed for this work will be made available upon acceptance
\footnote{Open Source Code: The Python code developed for this work will be made available upon acceptance}

% optionally more appendix sections here

%\section{Additional Figures} ...

\bibliographystyle{IEEEtran}
\bibliography{bib/IEEEabrv,bib/related_work,bib/force_intrinsic,bib/parallel_robots,bib/Upper_urinary_cancer,bib/force,bib/shape}

\end{document}